\documentclass [sigconf]{acmart}
\usepackage{fancyhdr}
\usepackage{amsmath}
\usepackage{wrapfig}
\usepackage{multirow}
\usepackage{graphicx}
\usepackage{subfigure}
\usepackage{makecell}
\usepackage{algorithm}  
\usepackage{graphicx}
\usepackage{multirow}
\usepackage{svg}
\usepackage{bm}
\usepackage{amsfonts}
\usepackage{algpseudocode}
\AtBeginDocument{%
  \providecommand\BibTeX{{%
    \normalfont B\kern-0.5em{\scshape i\kern-0.25em b}\kern-0.8em\TeX}}}

\copyrightyear{2024}
\acmYear{2024}
\setcopyright{acmlicensed}
\acmConference[WWW '24 Companion]{Companion
Proceedings of the ACM Web Conference 2024}{May 13--17, 2024}{Singapore,
Singapore}
\acmBooktitle{Companion Proceedings of the ACM Web Conference 2024 (WWW
'24 Companion), May 13--17, 2024, Singapore, Singapore}
\acmDOI{10.1145/3589335.3651243}
\acmISBN{979-8-4007-0172-6/24/05}
\begin{document}
% \settopmatter{printacmref=false}
% \settopmatter{printacmref=false,  printccs=true}
%renewcommand\footnotetextcopyrightpermission[1]{}
\title{Towards Personalized Evaluation of Large Language Models with An Anonymous Crowd-Sourcing Platform}

\author{Mingyue Cheng\textsuperscript{1}, Hao Zhang\textsuperscript{1}, Jiqian Yang\textsuperscript{1}, Qi Liu\textsuperscript{1*}, Li Li\textsuperscript{1}, Xin Huang\textsuperscript{1}, Liwei Song\textsuperscript{1},\\ Zhi Li\textsuperscript{2}, Zhenya Huang\textsuperscript{1}, Enhong Chen\textsuperscript{1}}\thanks{Qi Liu is corresponding author.}
\affiliation{%
	\institution{$^1$Anhui Province Key Laboratory of Big Data Analysis and Application, University of Science and Technology of China \& State Key Laboratory of Cognitive Intelligence, Hefei, China, \\ $^2$  Shenzhen International Graduate School, Tsinghua University, Shenzhen, China}
	\country{}}

\email{{mycheng, qiliuql, huangzhy, cheneh}@ustc.edu.cn,{zh2001,yangjq, lili0516, wuli\_error, songliv}@mail.ustc.edu.cn, 
}
\email{zhilizl@sz.tsinghua.edu.cn}

\begin{abstract}
Large language model evaluation plays a pivotal role in the enhancement of its capacity. Previously, numerous methods for evaluating large language models have been proposed in this area. Despite their effectiveness, these existing works mainly focus on assessing objective questions, overlooking the capability to evaluate subjective questions which is extremely common for large language models. Additionally, these methods predominantly utilize centralized datasets for evaluation, with question banks concentrated within the evaluation platforms themselves. Moreover, the evaluation processes employed by these platforms often overlook personalized factors, neglecting to consider the individual characteristics of both the evaluators and the models being evaluated. To address these limitations, we propose a novel anonymous crowd-sourcing evaluation platform, BingJian, for large language models that employs a competitive scoring mechanism where users participate in ranking models based on their performance. This platform stands out not only for its support of centralized evaluations to assess the general capabilities of models but also for offering an open evaluation gateway. Through this gateway, users have the opportunity to submit their questions, testing the models on a personalized and potentially broader range of capabilities. Furthermore, our platform introduces personalized evaluation scenarios, leveraging various forms of human-computer interaction to assess large language models in a manner that accounts for individual user preferences and contexts. The demonstration of BingJian can be accessed at \url{https://github.com/Mingyue-Cheng/Bingjian}. 
\end{abstract}
\keywords{Large Language Model, Personalized Evaluation, Crowdsourcing Platform}
\ccsdesc[500]{Computing methodologies~ Natural language
processing}
\maketitle
% \let\thefootnote\relax\footnotetext{Qi Liu is corresponding author.}

%
% \ccsdesc[500]{Information systems~Recommender systems}
% %\ccsdesc[300]{Computer systems organization~Redundancy}
% %\ccsdesc{Computer systems organization~Robotics}
% \ccsdesc[100]{Collaborative filtering}

\vspace{-0.2in}
\section{Introduction}
% Research Background
The advent of Large Language Models (LLMs) has marked a significant milestone in the journey toward Artificial General Intelligence (AGI) \cite{luo2023unlocking,jiang2023reformulating}, opening new frontiers in our ability to process and understand complex human languages at an unprecedented scale. As these models become increasingly sophisticated, their evaluation transcends traditional paradigms \cite{wang2018glue}, challenging the very notion of a singular, definitive ground truth. In the realm of large language model development, the absence of a clear-cut benchmark necessitates a reimagined approach to evaluation—one that accommodates the nuanced and multifaceted nature of tasks these models are designed to tackle. This shift underscores the critical role of evaluation methodologies in not only benchmarking current capabilities but also in driving the evolution of model sophistication. The quality and depth of these evaluation mechanisms, therefore, directly influence the trajectory of large language model advancements, making it imperative to explore and refine our evaluative frameworks to keep pace with the rapid advancements in this domain.

\begin{figure*}[t]
    \centering
    \vspace{-0.1in}
    \includegraphics[width=0.9\linewidth]
    {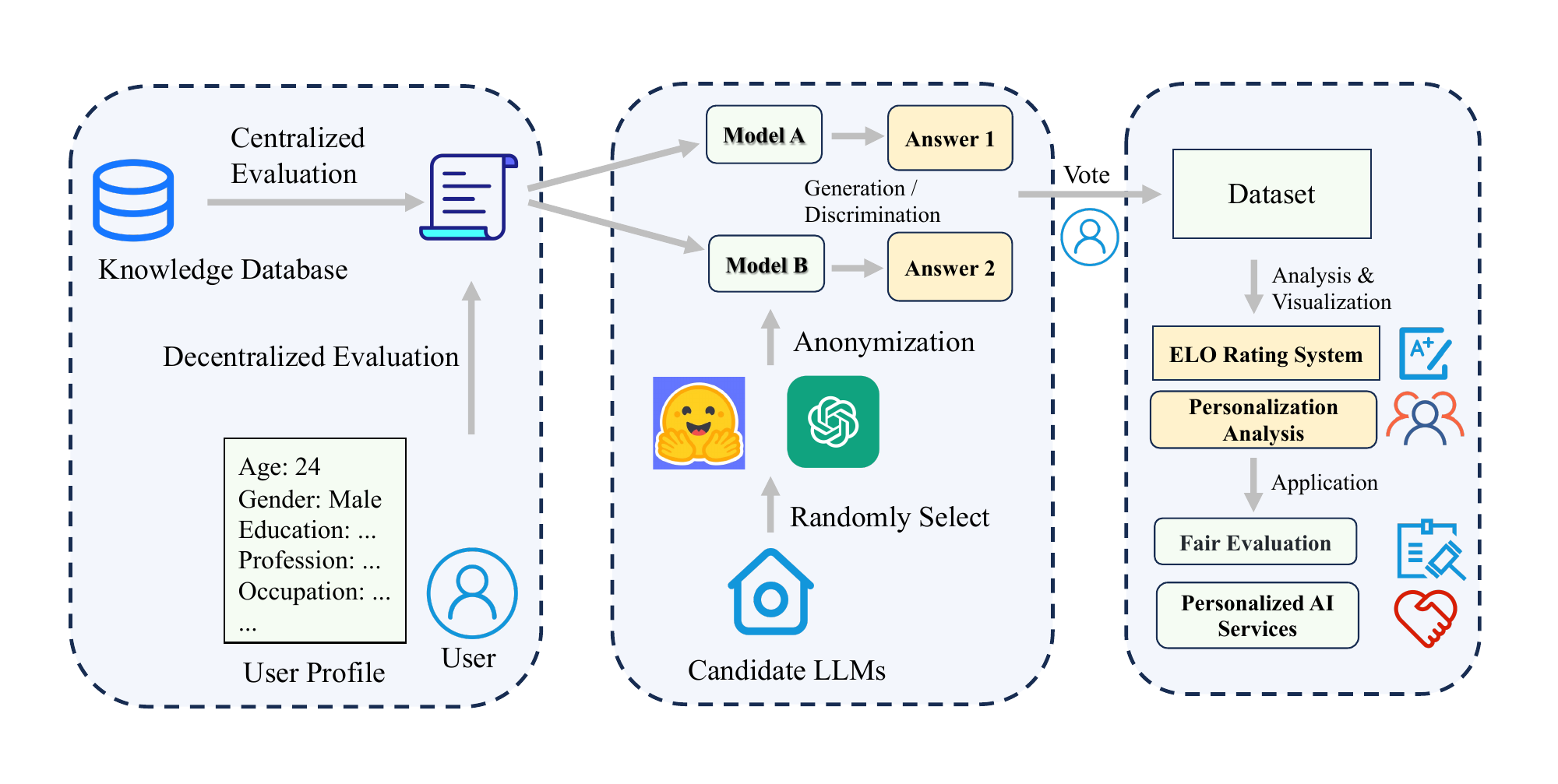}
    \vspace{-0.3in}
    \caption{The illustration of evaluation pipeline of BingJian platform.}
     \vspace{-0.1in}
    \label{BJ}
\end{figure*}
In the quest to evaluate LLMs, researchers are diligently working to gauge an expansive range of model capabilities, from coding proficiency to domain-specific expertise. These efforts \cite{chang2023survey,zhuang2023efficiently,li2023beyond} play a crucial role in refining evaluation methodologies and shedding light on the complex competencies of LLMs. Yet, in spite of these advances, current methods face significant shortcomings that demand attention. One major issue is the dependence on centralized datasets, which narrows the evaluation to a set of predetermined challenges and fails to encompass decentralized, real-world problems. Additionally, most evaluation frameworks do not adequately consider the integration of personalized user data \cite{ding2018improved,zhao2014leveraging}, an essential factor that could provide deeper insights into how models perform across varied user interactions. These challenges highlight the necessity for inventive evaluation approaches that not only broaden the scope of problem collection to include decentralized issues but also factor in individual user contexts, thereby making the evaluation results more relevant and applicable.

To address the prevailing challenges in the evaluation of LLMs, we design a platform named BingJian aimed at facilitating comprehensive model assessment. On this platform, responses from different models are presented to users, who then are encouraged to select the most appropriate answer. To ensure a fair evaluation of model capabilities, BingJian employs an ELO rating system \cite{pelanek2016applications} that adjusts model scores based on user selections. Furthermore, BingJian is designed as an open, crowdsourced platform. Just as ImageNet \cite{deng2009imagenet} significantly advanced the field of computer vision by constructing a high-quality image dataset 
through crowdsourced annotations, we aim to revolutionize the evaluation of large language models. Traditional objective assessments \cite{huang2023c} fall short of capturing the full capabilities of LLMs. By incorporating human crowdsourcing evaluations, we introduce the most authentic form of human feedback. Humans assess models based on various intangible aspects, such as the quality of generated text, knowledge conveyed, and presentation style, offering a comprehensive evaluation beyond quantifiable metrics. Naturally, a crowdsourced platform can gather a wide variety of evaluation results. These outcomes are closely related not only to objective facts but also to the personalized information of the evaluators. To this end, our platform compiles a comprehensive dataset of evaluator personalization information. We aim to delve into this personal data to uncover the cognitive relationships between humans and LLMs. This endeavor provides a more holistic depiction of the evaluation results for LLMs, revealing how the individual characteristics of evaluators influence their interactions with and assessments of these models. By exploring these nuanced relationships, we can enhance our understanding of model performance from a perspective that integrates human subjectivity, thereby enriching the evaluation process.

\vspace{-0.1in}
\section{The Proposed BingJian}
In this section, we introduce our BingJian platform, as depicted in Figure \ref{BJ}. Initially, we establish a comprehensive question database encompassing a wide array of domains, along with gathering a suite of large language models for assessment. Following this, we develop interfaces incorporating both centralized and decentralized approaches to evaluation, enabling the collection of varied data sets pivotal for evaluating the multifaceted capabilities of LLMs. Next, we provide a detailed introduction to the BingJian platform, focusing on the interface design, the evaluation process, and the data analysis and visualization.
\vspace{-0.05in}
\subsection{Interface Overview}
\begin{figure}[t]
    \centering
    \vspace{-0.1in}
    \includegraphics[width=1\linewidth,height=150pt]{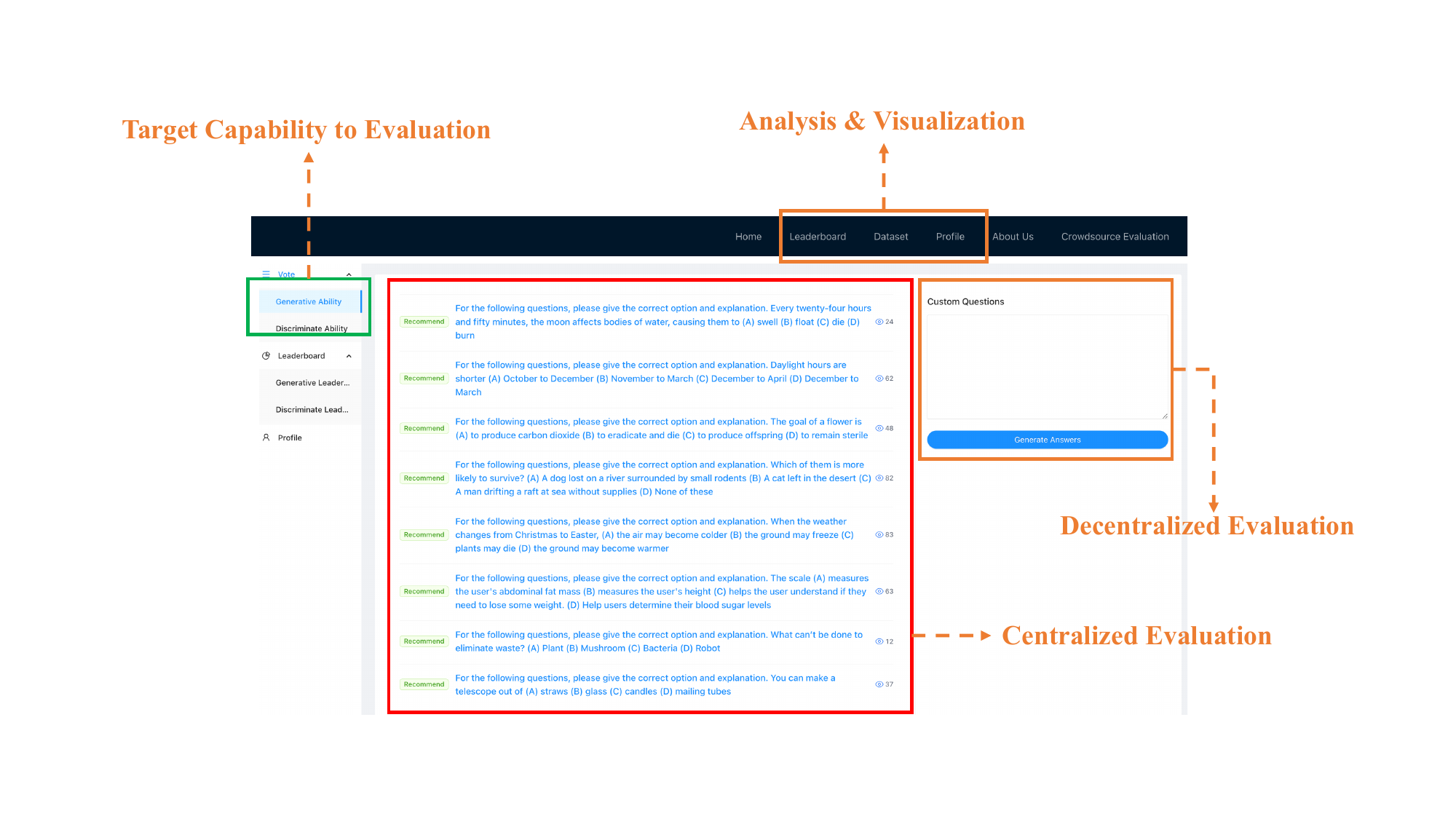}
    \vspace{-0.4in}
    \caption{Centralized and decentralized evaluation interface.}
    \label{overview}
    \vspace{-0.2in}
\end{figure}

% 如图1所示，冰鉴评测界面主要包括中心化评测和去中心化评测两部分设计。下面，我们分别详细介绍这两个模块在段2.1.1,2.1.2。
On the login page, we first encourage users to fill out their profile information, including age, gender, profession, and educational background, to facilitate subsequent analysis of personalized large language model evaluation results.
Then, as shown in Figure \ref{overview}, the BingJian evaluation interface primarily consists of two parts: centralized evaluation and decentralized evaluation.
\subsubsection{Centralized Evaluation}
The central evaluation is to evaluate the performance of LLM on a constructed question set, which covers general knowledge in various domains such as natural sciences, humanities, economics, etc. In each interaction, the users may choose a question displayed on the page. Then, the system will randomly invoke two different models to answer the question and anonymously display their answers along with the explanations and analyses, on the evaluation interface. Such an anonymous mechanism could help eliminate bias against different LLMs to some extent, effectively ensuring the fairness of the evaluation. 

Furthermore, we also integrate a question-recommendation system. Based on users' past question browsing history, we select questions that align with their interests to push to them while ensuring, as much as possible, that users do not encounter the same question twice. From the perspective of user experience, this approach significantly enhances the engaging nature of the crowdsourced evaluation process. When considered from an evaluation standpoint, this strategy aims to broaden users' evaluation activities across various related disciplines, thereby minimizing the risk of inaccurate assessments. This personalized and dynamic question recommendation not only caters to the users' preferences but also enriches the evaluation dataset by capturing a wider spectrum of user interactions and responses, leading to more robust and comprehensive insights into model performance.
\subsubsection{Decentralized Evaluation}
Due to the continuous iteration and updating of LLMs, the dataset in the constructed question bank may be incorporated into the model's training corpus in the present or future, leading to biases in the evaluation results. To address this challenge, we further design the decentralized evaluation module shown in the right column of Figure \ref{overview}. Users can input custom questions into the dialogue box and then click the button below to evaluate the generated answers. This feature greatly alleviates the potential issue of question leakage in the evaluation nowadays and further improves the fairness of the leader board while supporting the open-domain question.
\vspace{-0.1in}
\subsection{Crowdsource Evaluation}
% \subsubsection{Evaluation Process}
Just as ImageNet revolutionized computer vision research by involving a large number of contributors in labeling images, our platform aims to harness the power of crowdsourcing to establish a comprehensive evaluation process and database for large language models. 
% This approach also enables us to gather a diverse range of opinions and feedback, ensuring a broad spectrum of perspectives and reliable assessments. 
Next, we outline the specific evaluation process following the three primary data collection goals, respectively.
Through the implementation of crowdsourced evaluation, our objective is twofold: firstly, to objectively delineate the capabilities of numerous LLMs, and secondly, to foster the development of a benchmarking system through human-computer interaction assessments that will propel further advancements in LLM technologies. This benchmark will serve as a valuable resource for researchers and developers alike, offering a standardized framework against which the progression of model capabilities can be rigorously tested and compared.
\subsubsection{General Knowledge Mastery}
% 为评估模型的通用知识能力，我们构建了自然、科学、人文、经济等不同领域的问题库。所有的选择题用以下的prompt输入大模型。如图2所示，模型需要给出问题的正确选项。将其与正确答案匹配，则可统计模型的回答准确率从而分析模型的能力。
To evaluate the general knowledge mastery of the models, we have created question banks in various domains such as nature, science, the humanities, and economics. The multiple-choice questions will be presented to LLM with the following prompt:\textit{ \textbf{For the following questions, please give the correct option and explanation. <Question>, (A) <Answer1>, (B) <Answer2>, (C) <Answer3>, (D) <Answer4>.}} Then, the models are required to provide the correct options for the given questions. By matching their responses with the correct answers, we can calculate the accuracy of the model's answers and also preliminary conclude its capabilities over various domains.
\subsubsection{Generative Ability}
\begin{figure}[t]
    \centering
    \vspace{-0.1in}
    \includegraphics[width=1.1\linewidth,height=120pt]{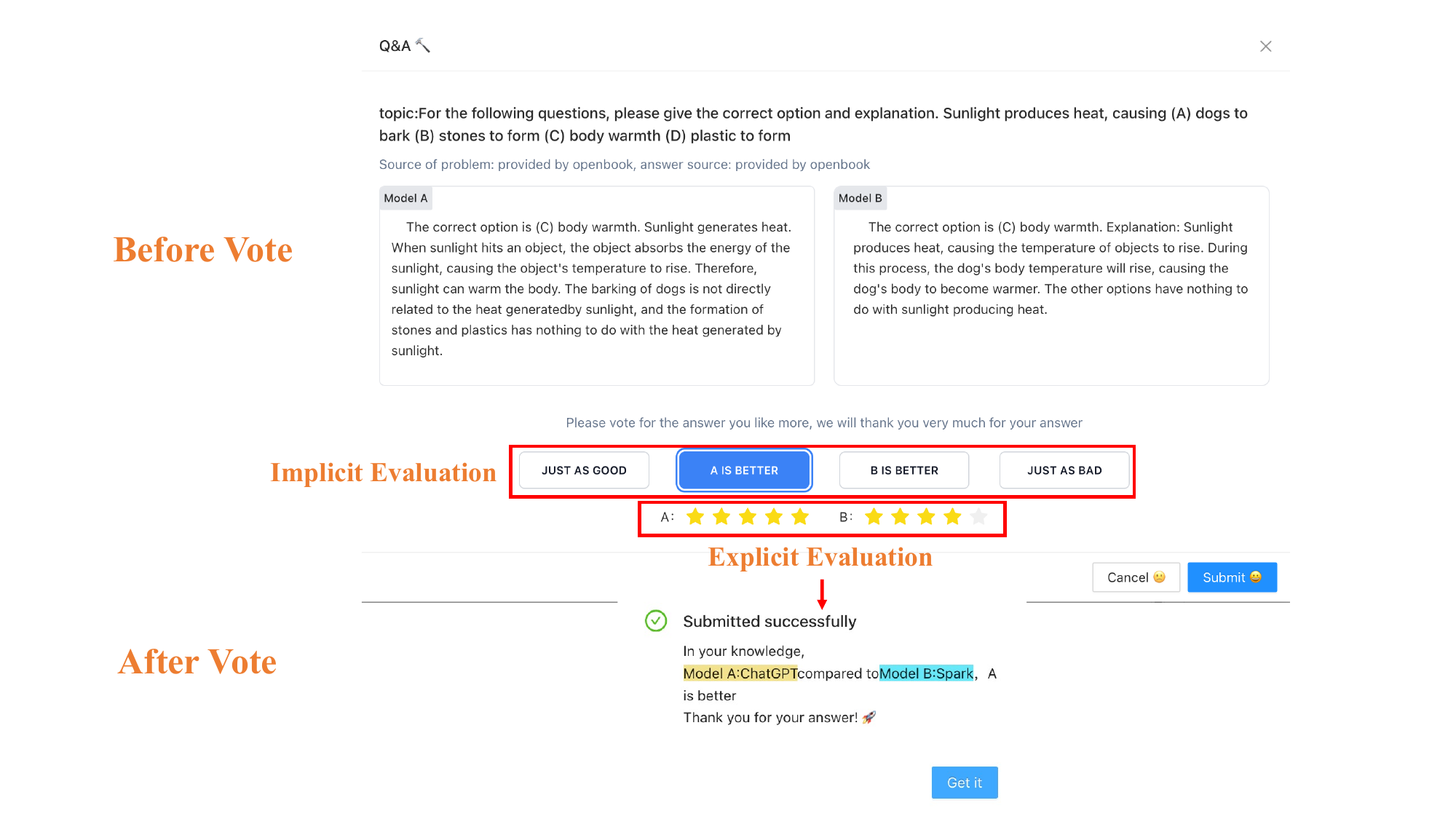}
    \caption{Evaluation process of generative ability.}
    \label{ge}
    \vspace{-0.1in}
\end{figure}
% 除了问题答案，答案的可解释性也是一项重要的评测指标。如图2所示，用户被鼓励综合模型给出的答案及其分析对模型的生成能力进行打分。
In addition to the problem answer, explainability is also an important evaluation indicator. As shown in Figure \ref{ge}, users are encouraged to score the model's generative ability based on the answers and their analysis.
The evaluation mainly involves a comparison (i.e., "JUST AS GOOD", "A IS BETTER", "B IS BETTER", "JUST AS BAD") of the outputs generated by different models. Besides, it also includes a quantitative evaluation of the model's generative capacity, as assessed using an exscoring system ranging from 1 to 5. Furthermore, given the substantial amount of user profile information collected, we highlight that this data can be utilized to analyze the correlation between the model's responses and personalized user profiles, providing potential research opportunities in the realm of personalized AI services.
For instance, certain user groups may be more inclined to prefer professional explanations, while others may favor imaginative responses.
\subsubsection{Discriminate Ability}
% 除了直接评测大模型直接生成的回答，我们还评测了大模型对不同的答案的判别结果，以分析大模型的判别能力。具体地，如图3所示，对于某个指定的问题，系统首先随机调用两个模型a，b并展示他们的回答
Evaluating the discriminate ability of large language models is essential to ensuring their reliability, addressing biases, and benchmark performance, providing valuable insights into their generalization capabilities in real-world applications. In this vein, our evaluation framework goes beyond simply assessing the generative prowess of large language models. We also scrutinize their ability to evaluate and judge different answers by engaging them in a comparative analysis. Specifically, for a given question, our system employs a double-blind method where it randomly selects two models, referred to as A and B, to provide answers without revealing their identities. These responses are then displayed on the user interface.Finally, users are invited to participate by scoring the responses given by models C and D to evaluate the discriminating ability of the large language models. Through this multi-tiered evaluation process, we can identify strengths and weaknesses in the models' abilities to discriminate between high- and low-quality responses. Such insights are instrumental for iterative improvements, leading to more sophisticated and reliable AI systems. Ultimately, by enhancing the discriminative capabilities of LLMs, we can better tailor them to a variety of applications, ensuring that they not only produce content that is engaging and informative but also critically sound and contextually appropriate.
% 然后，另外两个模型c，d将会被调用with 如下的prompt。。。。最终，用户需要对c，d的判别结果及理由进行评测。while大模型的判别能力作为重要的一部分，具有广泛的应用。这种评测的框架支持了对这种能力的分析，具有重要的现实意义
\vspace{-0.05in}
\subsection{Analysis \& Visualization}
\subsubsection{ELO Rating Mechaminsm}
Inspired by the renowned chess ranking system, the Elo Rating System (ELO) provides a dynamic and intuitive framework for gauging relative strengths. We initialize models with ELO ratings, in which winners gain ELO points while losers lose points, ensuring a continuous adjustment of model rankings based on relative model performance. 
To be specific, it updates a participant's rating ($R'$) based on the outcome ($S$) of each evaluation record, which can be expressed as:
\[ R' = R + K \cdot (S - E), \]
where $R'$ is the updated ELO rating, $R$ is the pre-match ELO rating, $K$ is a constant determining the rating change magnitude, $S$ is the match outcome (1 for a win, 0 for a loss, 0.5 for a draw) and  $E$ is the expected outcome calculated using a logistic function, i.e.
$ E = \frac{1}{1 + 10^{((R_B - R_A)/400)}}, $
in which $R_A$ and $R_B$ are the initial ELO ratings of the two participants. The logistic function ensures that the expected outcome aligns with the participants' relative ratings, making ELO a dynamic and reliable measure of skill in various competitive scenarios. This innovative methodology allows for a fair and nuanced assessment of the model's capability, transcending traditional evaluation metrics.
 \begin{figure}[t]
    \centering
    \includegraphics[width=1\linewidth]{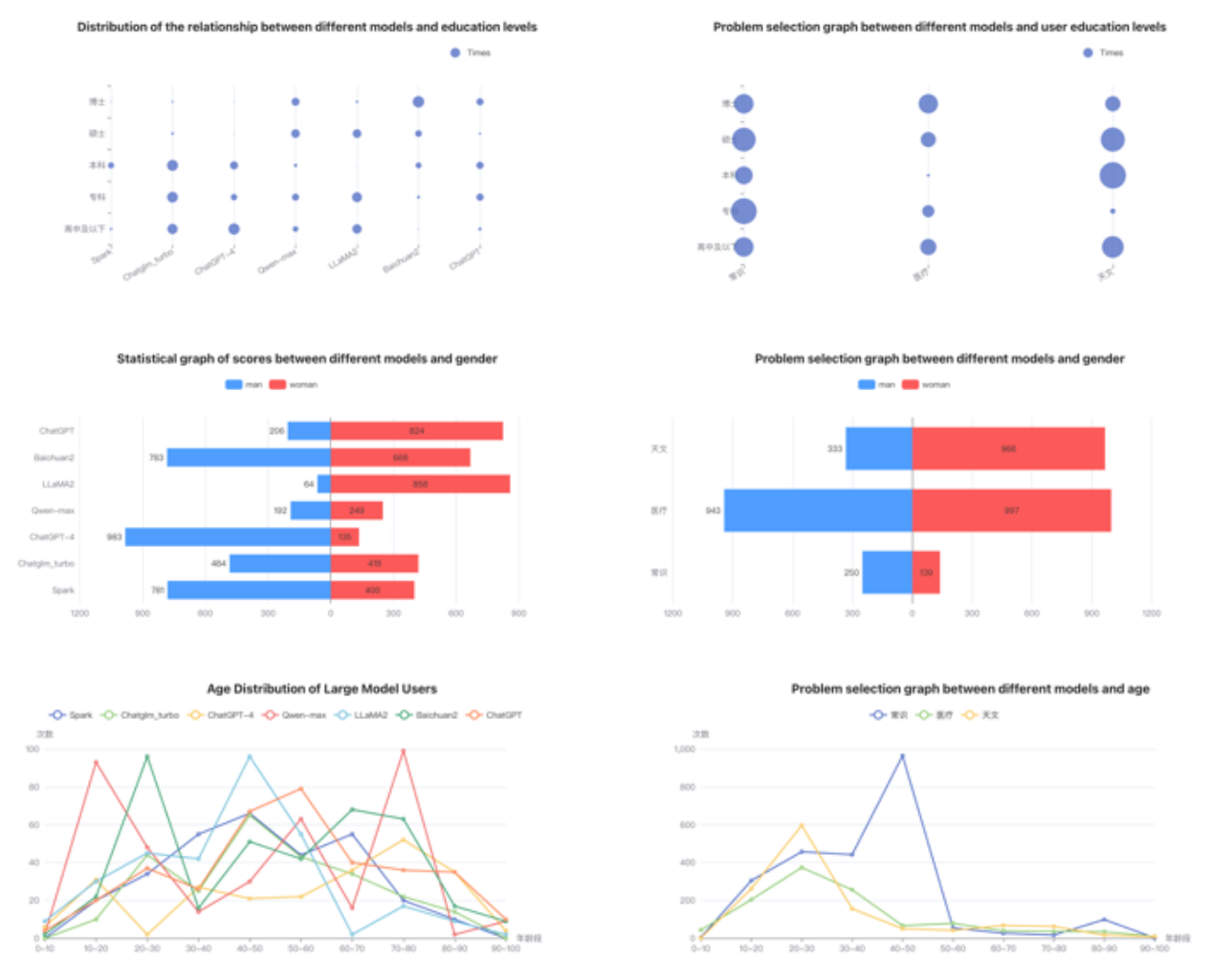}
    \vspace{-0.3in}
    \caption{Visualization of correlation between the model's responses and personalized user group profiles.}
    \label{v}
    \vspace{-0.2in}
\end{figure}
\subsubsection{Crowd Analysis}
To delve into the relationship between the responses generated by the model and the diverse backgrounds of the user groups, we embarked on a comprehensive visual analysis, as delineated in Figure \ref{v}. This analysis scrutinized the evaluation data through the lens of various demographic dimensions, such as age, gender, profession, occupation, and educational attainment. Our information collection requires user approval. Our objective in this exploratory endeavor is to detect and understand patterns that could inform and enhance the customization of services provided by large language models.
For example, we might discover that specific demographic segments have a predilection for responses that are steeped in professional jargon or technical detail, while others might demonstrate a preference for responses that are more creative or narrative in nature. Moreover, by examining the assessment results from these diverse demographic vantage points, we can identify unique opportunities for research into tailored AI services.
This granular analysis not only aids in refining the user experience but also serves as a foundational step towards the development of AI systems that are sensitive to the nuanced needs and preferences of different user groups. By integrating these insights into the iterative design of large language models, we can move closer to achieving a level of personalized interaction that mirrors the adaptive and discerning nature of human communication.
\section{Conclusion}
This paper introduced a personalized, anonymized crowd-sourcing platform for evaluating the capacity of large language models, providing users with both centralized and decentralized evaluation entry points. Users are enabled to assess models' generative and discriminative capabilities within this framework. Moreover, the platform conducts a comprehensive analysis of users' personalized information in conjunction with model evaluation results, utilizing visual statistical charts to display relevant profile information. 
% Currently, the platform has integrated a multitude of large language model interfaces and plans to further expand the number of models accessible for evaluation.
This innovative approach not only enriches the evaluation landscape by incorporating a human-centric perspective but also paves the way for a more nuanced understanding of model capabilities across diverse user backgrounds and preferences. The ongoing expansion of model integrations underscores our commitment to offer a robust and dynamic evaluation environment, poised to adapt and evolve with the advancing frontiers of large language model technologies.

\textbf{Acknowledgements}.
This research was supported by grants from the National Key Research and Development Program of China (Grant No. 2021YFF0901003), the National Natural Science Foundation of China (Grants No. 62337001, U20A20229), and the Fundamental Research Funds for the Central Universities. This work also thanks the support of funding of SC5290005194. We thank the Hefei Artificial Intelligence Computing Center of Hefei Big Data Asset Operation Co., Ltd. for providing computational resources for this project. 
\bibliographystyle{ACM-Reference-Format}
\bibliography{sample-base}
\end{document}